\documentclass[sigconf]{aamas}  

\usepackage{booktabs}

\acmDOI{doi}  
\acmYear{2020}  
\acmPrice{}  

\usepackage{amsmath}
\DeclareMathOperator*{\argmax}{\arg\!\max}

\DeclareMathOperator{\EX}{\mathbb{E}}

\usepackage{mathtools}
\usepackage[mathscr]{euscript}
\usepackage{graphicx}
\usepackage{tikz,graphics,color,fullpage,float,epsf,caption,subcaption}

\hypersetup{draft}
\begin{document}

\title{Adversarially Guided Self-Play for Adopting Social Conventions}  


\author{Mycal Tucker}
\affiliation{%
    \institution{MIT}
    \city{Cambridge}
    \state{Massachusetts}
}
\email{mycal@csail.mit.edu}

\author{Yilun Zhou}
\affiliation{%
    \institution{MIT}
    \city{Cambridge}
    \state{Massachusetts}
}
\email{yilun@csail.mit.edu}

\author{Julie Shah}
\affiliation{%
    \institution{MIT}
    \city{Cambridge}
    \state{Massachusetts}
}
\email{julie\_a\_shah@csail.mit.edu}

\begin{abstract}  
Robotic agents must adopt existing social conventions in order to be effective teammates.
These social conventions, such as driving on the right or left side of the road, are arbitrary choices among optimal policies, but all agents on a successful team must use the same convention.
Prior work has identified a method of combining self-play with paired input-output data gathered from existing agents in order to learn their social convention without interacting with them.
We build upon this work by introducing a technique called Adversarial Self-Play (ASP) that uses adversarial training to shape the space of possible learned policies and substantially improves learning efficiency.
ASP only requires the addition of unpaired data: a dataset of outputs produced by the social convention without associated inputs.
Theoretical analysis reveals how ASP shapes the policy space and the circumstances (when behaviors are clustered or exhibit some other structure) under which it offers the greatest benefits.
Empirical results across three domains confirm ASP's advantages: it produces models that more closely match the desired social convention when given as few as two paired datapoints.
\end{abstract}


\maketitle

\def\UrlBreaks{\do\/\do-}


\section{Introduction}

Humans routinely conform to social norms in order to successfully collaborate with others.
For example, drivers in France learn to drive on the right side of the road, while their counterparts in England drive on the left side, because doing so matches each respective community's social norm.
However, a French driver is capable of quickly understanding the change in convention when in England by leveraging their previous experience and observing other drivers within the new environment; in short, for humans, explicit instruction is helpful but not necessary.

If robots are to be capable teammates in multi-agent systems, they should behave similarly to a French driver in England by adopting the social norms or conventions of their partners.
In many ways, robots that interact with humans in daily life are already programmed to conform to conventions: voice assistants, for example, use language as a convention for communication, and self-driving cars drive on the correct side of the road.

While pre-programming agents to conform to fixed social conventions works in some cases, it necessarily fails to handle unforeseen scenarios.
Theoretically, learning techniques could address this gap, but most approaches to learning social conventions in multi-agent systems (such as policy cloning or interacting directly with other agents) require large amounts of data \cite{OSP}.many orders of magnitude of data \cite{OSP}.
However, if a French driver can switch to driving on the left side of the road after only a few glimpses of English traffic, autonomous agents should be able to make similar adjustments to their own behavior.

Prior work has introduced observationally augmented self-play (OSP), a technique that takes a step toward learning social conventions with limited data \cite{OSP}.
The authors' key insight is to combine self-play with paired data (inputs and outputs generated by agents conforming to a given social convention that can be used for supervised learning).

While the results presented by \citet{OSP} represent a significant advance in learning social conventions, gathering the paired data required by OSP may be difficult in the real world.
For the French driver mentioned above, paired data would take the form of all the factors English drivers consider and the actions they subsequently take.
Acquiring such data may be impossible, and humans do not find it necessary; instead, we desire that our agents learn in the same way that humans learn how to drive on English roads by merely observing that drivers persistently remain on the left side of the road regardless of their state.

In this work, we propose adversarial self-play (ASP), a new technique that uses unpaired data (i.e., only the outputs of agents using a social convention without associated inputs) in adversarial training in order to increase the likelihood of newly trained agents adopting the desired pre-existing social convention.
Our method incorporates a discriminator to distinguish between outputs produced by a training agent and those generated from the desired social convention while the training agent tries to fool the discriminator.
Combined with self-play and limited paired data, ASP consistently produces agents that better conform to social conventions compared with those produced by existing techniques.
Adversarial training of social conventions is a sufficiently general approach that ASP is well-suited for a variety of domains regardless of the number of agents, exact task, or learning technique.

We demonstrate the effectiveness of our approach in three applications: a temporally-extended speaker-listener domain, an autoencoder, and a multi-agent coordination game involving multiple communication steps.
Our findings indicate that ASP enables improved behavior across all settings, even when provided with only two or three examples of paired data.

\section{Preliminaries}
We wish to train a new agent such that, when the agent is introduced into a team, the entire team performs well.
The team's task may vary widely: perhaps a speaker must direct a listener to a target location, or perhaps multiple agents must coordinate to pull different levers.
In many cases, the tasks we are interested in may be formulated as multi-agent Markov decision processes (MDPs) in which the optimal actions of all agents depend only upon the current state.
Our goal, therefore, is for a new agent to learn a policy that maximizes the overall mixed-team reward (defined more formally below) in a joint MDP.

\subsection{Learning Policies for MDPs}
A (single agent) MDP is represented by the tuple of $\langle S, \allowbreak A, \allowbreak T, \allowbreak \allowbreak R, \allowbreak \gamma\rangle$, where $S$ is the state space, $A$ is the action space, $T: S\times A\rightarrow \mathbb P_S$ is the transition function that maps the current state and action to a distribution over next state, and $R: S\times A\times S\rightarrow \mathbb R$ is the reward function that produces a real-valued reward \cite{suttonbarto}.
The reward function can depend upon the current state, the current action, and the next state, even if it only depends on a subset of these in most problems.
$\gamma\in[0, 1]$ is the discount factor representing how much current reward is favored over future reward. 

The goal of solving an MDP is to identify a policy $\pi: S\rightarrow A$ that maximizes $\mathbb E_\pi\left[\sum_{t=1}^\infty \gamma^{t-1}r_t\right]$, where $r_t$ is calculated according to reward function $R$ based on state action trajectory $(s_1, a_1, s_2, a_2, ...)$ as determined by transition function $T$ and policy $\pi$. 

With $N$ agents, each agent $i$ has its own action set, $A_i$, and the transition function $T: S\times \prod_{i=1}^N A_i\rightarrow \mathbb P_S$ takes all agent actions and produces a distribution over the next state (assuming instantaneous actions). Each agent may have its own reward function $R_i$, which can also depend upon the joint actions. The goal, then, is for each agent to learn its own policy $\pi_i: S\rightarrow A_i$ in order to maximize individual reward. In our case, because we are working in cooperative settings, we use a single reward function $R$ that all agents share. 

\subsection{Mixed Team Rewards}
In the previous section, we framed the problem of learning optimal joint policies for an MDP.
This work, however, focuses on how mixed teams - teams composed of agents that have not been trained together - perform a multi-agent task.

Consider a team of size $N$ composed of $i$ instances of an agent $a$, each agent executing policy $\pi_a$, and $j = N - i$ instances of an agent $b$ executing policy $\pi_b$.
The joint policy of the team can be expressed as $\pi_{(a, i), (b, j)}$ with an associated expected discounted reward calculated similarly to any joint policy: $R(\pi_{(a, i), (b, j)}) = \mathbb E_{\pi_{(a, i), (b, j)}}\left[\sum_{t=1}^\infty \gamma^{t-1}r_t\right]$.

The previous expression describes the expected discounted reward for a particular mixed team; we wish, however, to measure the expected performance of all mixed teams, calculated as follows:

\begin{equation}
\bar R_m = \frac{1}{N - 1} \sum_{i=1}^{N-1} R(\pi_{(a, i), (b, N-i)})
\label{eq:rm}
\end{equation}

In this work, we assume a fixed ``base'' agent with policy $\pi_b$ and wish to identify a policy $\pi_a$ that maximizes the mixed team reward.
We instantiate $\pi_a$ as a neural net with parameters $\theta$, so the task of learning the optimal policy becomes one of learning settings for $\theta$.
Solving for optimal $\theta^*$ can be accomplished directly by training in mixed teams:

\begin{equation}
\theta^* = \argmax_{\theta \in \Theta} \EX_{\pi}\left[\sum_{t=1}^\infty \gamma^{t-1}\bar R_{m, t}\right]
\end{equation}

Numerous methods exist for training neural nets to maximize such functions (e.g. those outlined in \cite{marl0, maddpg, thesis}, among others), but they often require large amounts of data and, therefore extensive queries of how the base agent would behave in a given state.
In other words, if other agents are treated as part of the environment, training a new agent requires that all agents participate in each training episode.

The French driver in our earlier example learned to drive on the left side of the road without explicit coaching and without crashing into English drivers; autonomous agents should learn in a similar manner.
Therefore, we combine findings from two fields of existing research in order to yield more data-efficient learning of social conventions.

\section{Related Work}
\subsection{Adversarial Training}
ASP incorporates insights gleaned from previous uses of adversarial techniques in neural nets but applies them, to the best of the authors' knowledge, to learning social conventions for the first time.

In one branch of adversarial training research, generative adversarial networks (GANs) are now used to produce lifelike images from random noise \cite{gan, nvidia}.
In a GAN trained to generate photorealistic images of people, for example, a generator network produces candidate images that an adversary (or discriminator) then attempts to differentiate from photos of real people.

Moving beyond GANs generating images from random noise, multiple extensions demonstrate the power of adversarial training in image translation tasks.
The \texttt{pix2pix} toolkit uses conditional adversarial networks to convert images from one type to another (e.g. to transform a photo taken during the day to a nighttime photo) \cite{pix2pix}.
Similarly, CycleGAN uses discriminator losses and a cycle-consistency loss to transform images between classes with zero paired data.
For example, using a set of photos of zebras and a separate set of photos of horses, CycleGAN can produce the ``horse version'' of a zebra photo, and vice versa \cite{cyclegan}.
Although such applications are often rooted in image processing, they indicate that adversarial training can be used to learn complex functions with little or no labeled data.

One potential weakness of adversarial training is its inability to discriminate between policies that produce the same marginal distribution over outputs.
For example, in CycleGAN, if horses and zebras each face left 50\% of the time and right 50\% of the time, the GAN could learn to map left-facing horses to right-facing zebras and vice versa.
However, such permutations do not appear to occur often in practice: in CycleGAN, the fake zebras resemble the input horses with stripes painted on top.

Outside the image domain, and independently from GANs but still leveraging adversarial techniques, co-training agents in adversarial, zero-sum games has produced agents that can, for example, play Go better than humans do \cite{go}.
Pitting two neural nets against each other encourages policy exploration until an optimal strategy is found.

We focus on cooperative games in our work, but apply the broad insight that adversarial training encourages exploration and prunes away large policy spaces in order to shape how cooperative agents behave.
Training in collaborative games often converges to locally sub-optimal strategies without further exploration; we aim to use the advantages of adversarial training to overcome this weakness \cite{hanabi}.

\subsection{Social Conventions}

Social conventions (strategies employed by multi-agent systems in cooperative settings) have been observed among both humans and autonomous agents, and have therefore been studied in the fields of cognitive science and artificial intelligence.

Within cognitive science, researchers have studied the emergence of conventions both among small teams and within society more broadly \cite{emerge2}.
In partnerships working on referring tasks (wherein one partner describes an object that the other partner must identify), partners collaborate to form conventions \cite{referring, conceptualpact}.
Such conventions may emerge quite rapidly and based on little information: for example, when discussing New York City landmarks, partners quickly establish each other's levels of expertise and modify their speech accordingly \cite{nyc}.
Likewise, \citet{tangrams} found that strangers developed efficient communication patterns when repeatedly referring to objects.
After the use of initially uncertain language, pairs begin to shorten the length of their expressions, effectively compressing their communication into a more efficient but less generalizable convention.

Complementing studies of how humans form or adopt social conventions, research into multi-agent artificial intelligence teams has assessed how social conventions among autonomous agents are created and what forms they take.
Drawing directly upon referring-task literature that studies humans, numerous techniques have been developed for training referring agents that work well with humans; however, such work often uses hand-coded, pragmatic models to generate references \cite{context, unambig}.
In other words, the models do not learn an existing social convention but exhibit behavior designed ahead of time to match peoples' preferences.

In conjunction with human-facing agents, other researchers have studied how autonomous agents develop conventions among themselves \cite{learningconvs}.
While many multi-agent training techniques do not explicitly formulate their problems as related to forming social conventions, their agents nevertheless learn to cooperate and communicate effectively \cite{commnet, learningcomm, who,emerge, emergent}.
Occasionally, authors have attempted to interpret the learned social conventions (e.g. \citet{commnet}), but the emphasis of this work has primarily been on having a social convention emerge at all.
We, however, focus on adopting a particular existing social convention.

Lastly, and most directly related to this work, some recent research has explored how autonomous agents may learn and adopt existing social conventions \cite{OSP, friends, adhoc}.
Broadly, such techniques depend upon paired data or interactions with other agents; however, because gathering such data may be difficult, this is exactly the type of data we seek to use as little as possible in ASP.
As our work is most closely related to that presented in \citet{OSP}, we explain their approach in greater detail here.

\citet{OSP} introduced observationally augmented self-play (OSP), a technique for combining self-play with paired data, $P$, which takes the regular form of supervised-learning data (input-output pairs).
The addition of such data grows the ``basin of attraction,'' or the set of initial conditions that will lead to learning the social convention through self-play.
Furthermore, supplementing self-play training by initializing a model trained on such data can only improve the likelihood of learning the desired policy.
(Even just one data point in $P$ may help, but it can never hurt.)
Although the theoretical arguments for the advantages of $P$ hold when $P$ is only used in model initialization, in practice the authors interleave supervised training throughout.

While OSP produces impressive results, it requires \textit{paired} data: the dataset $P$ must consist of input-output pairs that describe how the base agent behaves in a particular state.
Unfortunately, paired data may be difficult or impossible to gather; in our earlier French driver example, inputs would have to span everything from views in the rear-view mirror to the mental workload of English drivers.
This seems difficult to obtain and probably unnecessary for learning to stay on the left side of the road.

\section{Technical Approach}
\label{sec:tech}

\subsection{Development of Approach}
Unlike paired data, unpaired datasets $U$ are often easier to gather.
For example, our French driver could simply observe that English drivers take actions consistent with driving on the left side of the road, without bothering to record what caused such behavior.
The problem we seek to solve, therefore, is learning a social convention using limited paired data ($P$), and we aim to do so by uncovering patterns in unpaired data ($U$).

As indicated earlier, adversarial techniques lend themselves quite naturally to scenarios involving large amounts of unpaired data.
Therefore, we build a loss function that augments self-play and supervised learning terms (as introduced in OSP) with a third term for adversarial training with unpaired data.
Adversaries are trained to predict, given an element of dataset $U$ and an output of the training model $u'$, which entry has been produced by the model.
Conversely, the model attempts to fool the adversary.

Thus, we optimize the parameters, $\theta$, of our model via gradient descent along a loss function composed of the weighted sum of three terms representing self-play, paired data, and unpaired data:

\begin{equation}
L^{ASP}(\theta, P, U) = L^{sp}(\theta) + \lambda_0 L^{p}(\theta, P) + \lambda_1 L^{u}(\theta, U)
\end{equation}

In our experiments, the paired loss is mean squared error (MSE), while the adversary loss function is negative binary cross-entropy.
We use the negative of the cross-entropy to encourage the model to fool the adversary, and the adversary is trained to minimize positive binary cross-entropy when classifying fake or real outputs.

While expected performance is non-decreasing as the size of $P$ increases, the relationship between $U$ and performance is more complex.

Assuming a sufficiently expressive adversary, adversarial training drives the main model to produce outputs that conform to the distribution of the unpaired data $U$.
This is the danger of adversarial training: the adversary will force ASP to learn a different social convention if $U$ does not properly approximate the true social convention's distribution.
In an extreme example, if $U$ comprises only a single datapoint, the adversary will try to force ASP to only use that single action, even if doing so results in sub-optimal self-play.
Thus, it is critical for the $U$ dataset to be large enough to properly capture the desired distribution.

\subsection{Analysis of Policy Space}
Assuming that $U$ matches the social convention's distribution, adversarial techniques offer benefits by shrinking, or pruning, the set of acceptable policies.

Consider a single-step discrete MDP with $S = |\mathscr{S}|$ discrete states, $A$ discrete actions, and a pre-trained and deterministic social policy $\pi_{SC}$.
We further assume that all of the initial states are equally likely.
Without additional information about how the policy behaves or the reward structure of the environment, $A^S$ possible policies exist, as the policy could take any of $A$ actions for each state.

If, however, $\pi_{SC}$ produces actions according to a distribution where $p_i = \sum_{s \in \mathscr{S} s.t. \pi_{SC}(s) = a_i} P(s)$, the set of possible policies shrinks significantly.
Specifically, policies must now belong to the set of all policies for which subsets of the state space with state probabilities that sum to $p_i$ map to action $a_i$.

A large number of such policies still exists, and dividing the state space into groups of states that all result in the same action is a combinatorics problem.
Thus, we conclude that there are ${S \choose p_0S, p_1S, ..., p_{A-1}S}$,the multinomial coefficient, policies allowed by adversarial training.

Two important implications arise from this conclusion:
\begin{enumerate}
	\item The advantage of adversarial training depends upon the distribution being mimicked. The multinomial coefficient has the well-known property of being maximized for a uniform distribution and minimized for a delta function \cite{multinomial}.
	\item Even in the worst case of a uniform distribution of actions, the set of possible policies decreases from $A^S$, as demonstrated below.
\end{enumerate}

We measure the degree of pruning (how many policies are eliminated) through adversarial training by evaluating the worst case, wherein all actions are equally likely.
Specifically, we examine the log ratio $lr$ of adversarially allowed policies and the set of all possible policies.

\begin{align}
	\begin{split}
		lr &= \ln\left({\frac{{S \choose p_0S,...p_{A-1}S}}{A^S}}\right)\label{eq:1}
	\end{split}\\
	\begin{split}
		& = \ln\left(\frac{S!}{{(\frac{S}{A}!)}^A}\right) - S\ln(A)
		\label{eq:3}
	\end{split}\\
	\begin{split}
		& = \ln(S!) - A\ln\left(\frac{S}{A}!\right) - S\ln(A)
		\label{eq:4}
	\end{split}\\
	\begin{split}
		& = S\ln(S) - S + O(\ln(S)) - S\ln\left(\frac{S}{A}\right) + S\\&\qquad \qquad - AO\left(\ln\left(\frac{S}{A}\right)\right) - S\ln(A)
		\label{eq:5}
	\end{split}\\
	\begin{split}
		& = O\left(\ln(S) - A\ln\left(\frac{S}{A}\right)\right)
		\label{eq:6}
	\end{split}
\end{align}

In Equation~\ref{eq:1}, we define the log ratio as the log of the number of policies permitted by ASP divided by the number of all policies.
Equations~\ref{eq:3} and \ref{eq:4} follow from the application of log rules and substituting the factorial form of the multinomial coefficient.
We apply Stirling's approximation (which remains accurate even for small values) to transform the factorials into a more familiar form in Equation~\ref{eq:5}, leading to the cancellation of terms \cite{stirling}.
Lastly, further application of log rules and cancellations results in our final form, presented in big-O notation.

Analyzing the final expression yields insight into ASP's performance.
First, the log ratio decreases as $S$ increases;
conversely, as $A$ increases, to the upper limit of $A = S$ (corresponding to a unique action for each state) the log ratio increases.
Thus, ASP's advantage is maximized for big state spaces and a small set of actions, while its effect is muted as the number of actions reaches parity with the number of states.

The above analysis considers only the worst case for ASP: a uniform distribution over actions.
In other scenarios, as the probability distribution of actions concentrates over fewer actions, the combinatorial term decreases, magnifying the benefits of ASP.

\section{Results}
\label{sec:results}
Here, we demonstrate the benefits of ASP in experiments across three applications.
The idea of using an adversary to shape the policy space is independent of the task itself or the learning technique being used for self-play.
Thus, we test ASP in RL and autoencoder settings with different learning algorithms and numbers of agents.

In our first domain, a speaker and actor net must coordinate over multiple timesteps as the speaker (which observes the entire world) directs the actor (which only observes the speaker's commands) to move to a fixed target location.
Mixed human-robot teams often must coordinate in such a manner; as such, this domain suggests how a robot could learn a human's communication convention.
In the second setting, an MNIST variational autoencoder, a new VAE must learn the same encoding and decoding functions as a pre-trained net; thus, representation itself becomes the social convention to adopt.
Finally, our third domain tests three-agent teams in a multi-step coordination game involving multiple communication conventions.

For all domains, we measured the task-specific performance of mixed teams (teams composed of copies of a pre-trained base agent that already exhibits a social convention and a new agent trained using ASP).
This metric corresponds with the expected mixed team reward introduced in Equation~\ref{eq:rm}.
Given the high variance in performance possible due to complex team interactions, we report median values along with first and third quartiles\footnote{Analysis of the means exhibited similar trends.}.

For each trial across all experiments, we trained a new base model before training new models to learn its convention, mitigating the observed effect of some models serving as better or worse partners in mixed teams.
Within a given trial, we trained new models using ASP and two other techniques: a policy cloning (PC) strategy that trains exclusively on paired examples without self-play and OSP.
As previously reported, we expected PC to fail in high dimensional domains and expected OSP to perform well with sufficient paired data \cite{OSP}.
By using unpaired data, we intended to demonstrate that ASP better learns social conventions with the same amount of paired data as OSP.

(Accompanying code will be released online after the anonymous portion of peer review.
All code is written in Python, using Keras with a tensorflow backend \cite{keras}.)

\subsection{Particle World}

\begin{figure}[ht]
	\centering
	\includegraphics[width=\linewidth, trim={0cm 2cm 0cm 0cm}, clip]{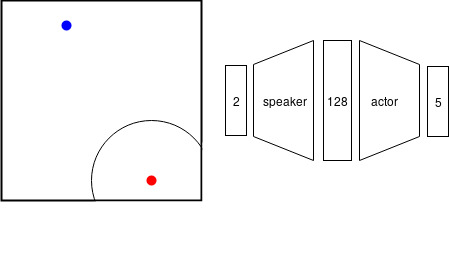}
	\caption{An actor in blue and a target in red spawn in random locations. The speaker observes the world and communicates to the actor, which then chooses an action to move within some radius of the target.}
	\label{fig:particles_world}
\end{figure}

We tested ASP in a two-agent particle world inspired by those used in \citet{MordatchA17}, \citet{maddpg}, and \citet{OSP}.
However, for simplicity, we only spawned a single target in a randomly chosen location, which remained fixed for the course of training and testing for a given trial, as shown in Figure~\ref{fig:particles_world}.
Because the target stayed fixed, its location was never provided: the pair implicitly learned where the target was by following high reward.
During an episode, which spanned 40 timesteps, a speaker agent received a two-dimensional state representation of the actor agent's $x$ and $y$ location.
At each timestep, the speaker agent produced a 128-dimensional communication vector, which was provided to the listener agent, yielding a softmax distribution over five possible actions, corresponding to moving one unit in any of the cardinal directions or remaining in place.
Reward was calculated as the inverse Euclidean distance from the actor to the target, unless the actor was within 1.4 units of the target, at which point the reward was set to 10.
The speaker and the listener shared the reward.
When trained in self-play with A2C, the team regularly achieved an average reward of 350 during a 40 timestep trial.
If a base model did not achieve a score of over 300 during self play, we discarded it and trained a new base.

The social convention to be learned was 1) what communication vector the speaker should produce given the state and 2) what action the actor should take given the communication vector.
We trained a speaker-actor pair to near-optimal behavior on its own; the subsequent task was to then train a new speaker-actor pair (speaker$'$ and actor$'$) such that, when speaker$'$ was paired with the base actor or the base speaker was paired with actor$'$, the mixed pair achieved high reward.

\begin{figure}[ht]
	\centering
	\includegraphics[width=\linewidth, trim={0.2cm 0.5cm 0.2cm 0.1cm}, clip]{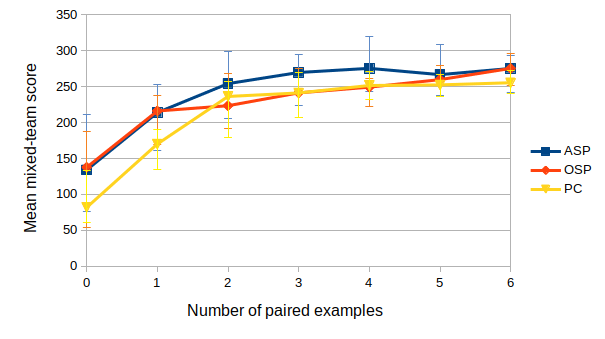}
	\caption{Agents trained with ASP performed better in mixed teams once paired data resolved initial ambiguity. Medians and quartiles plotted.}
	\label{fig:particles_graph}
\end{figure}

We tested ASP, OSP, and PC over 30 trials with the number paired examples of state-communication data ranging from 0 to 6.
ASP was given 4,096 unpaired data examples of communication vectors that the speaker produced.
Both paired and unpaired data were generated by running the desired number of trials and sampling uniformly at random from the 40 timesteps per trial.
Training consisted of 25,000 batches of size 8 sampled uniformly from the experience replay buffer, plus the full set of paired examples and an equal number of unpaired samples.
We measured mixed-team performance for a given trial by running 1000 trials with the base speaker and the new actor, and 1,000 trials of the new speaker with the base actor, and then recorded the mean score; the results are depicted in Figure~\ref{fig:particles_graph}.
Much of the variance in the graph is attributable to the different base models used for each trial.
We employed the non-parametric Friedman test on the data for trials with two, three, and four paired examples, viewing ASP, OSP, and PC as different ``treatments,'' and observed significant pairwise differences between ASP and both OSP and PC ($p < 0.01$ for each).

The advantage of ASP is most apparent in the presence of paired data.
With only zero or one paired data, the adversary alone could not resolve mapping ambiguity for which states should cause specific communication vectors.
As more paired data were introduced, the adversary allowed speaker$'$ to ``snap'' to the correct policy faster than either OSP or PC.
All three techniques benefited from additional paired data.

ASP derives its advantage from two sources.
First, the adversarial training prunes away large sections of the high-dimensional communication vector space.
Second, as noted in Section~\ref{sec:tech}, ASP benefits most from non-uniform probability distributions.
Given that the target location is selected at random in the 2D world, it is likely that the target is off-center; therefore, the distribution of communication vectors produced is unlikely to be uniform.
For example, if the target is located near the bottom of the world, it is reasonable to expect very few messages corresponding to ``move up'' commands.

\subsection{Autoencoder}

Adversarial shaping of social conventions is not limited to RL settings.
We evaluated ASP in a classic representation learning problem - an MNIST Variational Autoencoder (VAE) - to demonstrate how an adversary can shape learned latent representations to match the representation of an independently-trained VAE.
In order for the encoder of a separately trained VAE to work well with the decoder of another, it must learn the social convention of mappings from images to encodings.

\begin{figure}[ht]
	\centering
	\begin{subfigure}[b]{0.15\textwidth}
		\captionsetup{font=scriptsize}
		\includegraphics[width=\textwidth, trim={3.2cm 2.1cm 2cm 2.5cm}, clip]{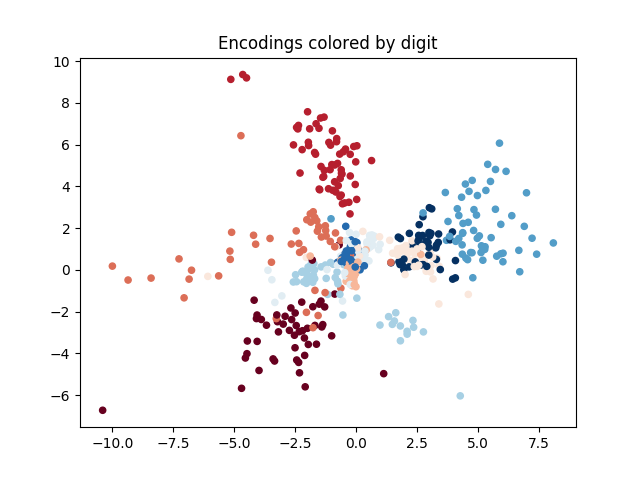}
		\caption{Base model}
	\end{subfigure}
	~ 
	\begin{subfigure}[b]{0.15\textwidth}
		\captionsetup{font=scriptsize}
		\includegraphics[width=\textwidth, trim={2.3cm 2.1cm 2cm 2.5cm}, clip]{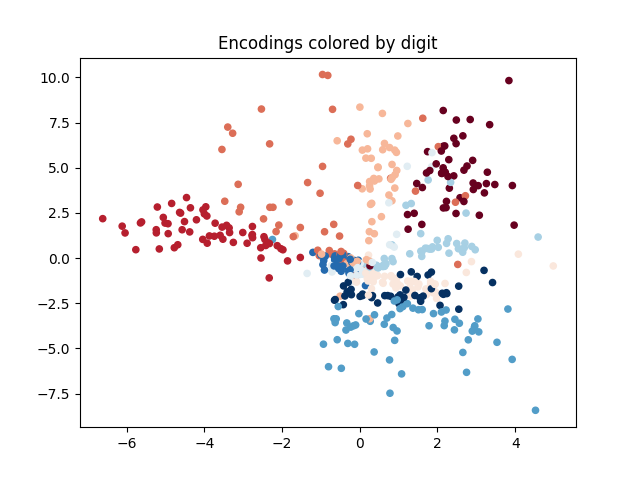}
		\caption{ASP + 0 paired}
	\end{subfigure}
	~
	\begin{subfigure}[b]{0.15\textwidth}
		\captionsetup{font=scriptsize}
		\includegraphics[width=\textwidth, trim={3.2cm 2.1cm 2cm 2.5cm}, clip]{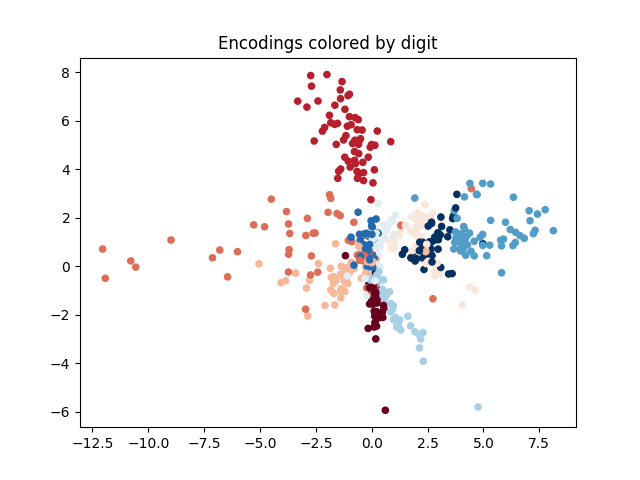}
		\caption{ASP + 8 paired}
	\end{subfigure}
	\caption{a) The base VAE clusters encodings by digit. ASP yielded a similarly-shaped latent space but for the wrong digits (b) until provided with eight paired examples (c).}\label{fig:mnist_latents}
\end{figure}

MNIST VAEs often learn somewhat interpretable representations that cluster images of the same digit near each other (Figure~\ref{fig:mnist_latents}) \cite{VAE}.
Plotting the neuron activations using a two-dimensional latent space and coloring by digit demonstrates both the benefits and limitations of ASP:
on one hand, even without paired data, ASP effectively guides a new model to use a similarly-shaped latent space as the base model's;
on the other hand, until paired data are introduced, the actual meaning of such points remains ambiguous.

We measured how well a VAE learned the social convention of another net by wiring the representation produced by one VAE's encoder into the decoder of the other VAE and calculating the mean squared error between the true MNIST input and its reconstruction.
Paired data are input image, encoding, and reconstructed image tuples, where the first two elements were used to train the encoder and the last two were used to train the decoder.
Unpaired data consisted of a set of 2,048 encodings.

It is worth noting that, in VAEs, the traditional training loss function comprises one reconstruction loss term (often the mean squared error - henceforth MSE - between the input image and the reconstructed image) and a regularization term that penalizes the net if the distribution of points in the latent space diverges from a unit Gaussian.
This second term acts similarly to the adversary in ASP: it constrains the distribution without specifying the exact mapping from inputs into the encoding space.
Therefore, when using ASP with a VAE, we must examine whether the adversary has any effect at all or if the latent distribution always matches a Gaussian.

We trained 10 VAEs over 5,000 batches of size 128 using the full MNIST dataset.
Those 10 models were each used in a single trial as the base model with social conventions that the new VAEs had to learn.

\begin{figure}[ht]
	\centering
	\includegraphics[width=\linewidth, trim={0.2cm 0.0cm 0.2cm 0.1cm}, clip]{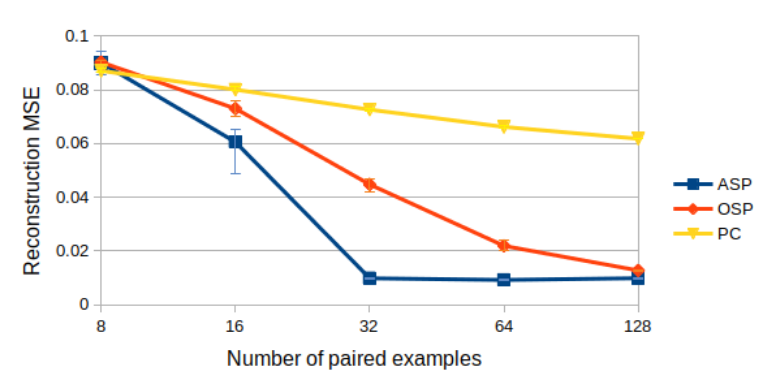}
	\caption{At extremes, OSP and ASP performed similarly, but ASP produced better reconstructions for intermediate amounts of paired data.}
	\label{fig:mnist0}
\end{figure}

In our first set of experiments, we used ASP, OSP, and PC for a VAE with latent dimension 32 and depicted in Figure~\ref{fig:mnist0} how each mixed model's reconstruction error changed as the number of paired examples increased.
Mixed models were composed of either the base encoder and the new decoder, or the new encoder and the base decoder.

As expected, all three models benefited from a larger amount of paired data.
In interpreting the plotted MSE values, a useful rule of thumb is that an MSE of 0.06 roughly corresponded to reconstructing the right digit, while an MSE of 0.006 (achieved during self-play with this experiment's settings) reproduced style as well.
Starting with just 16 examples, ASP began to generate reconstructions that were better than merely getting the digit right and that with 32 examples, reconstructions are nearly as good as those generated by a VAE trained entirely on its own.
More broadly, the same pattern of diverging and converging ASP and OSP played out in MNIST and the particle world: ASP offered the most benefit when sufficient paired data were provided to resolve mapping ambiguity but not so much that all behavior was prescribed.

\begin{figure}[ht]
	\centering
	\includegraphics[width=\linewidth, trim={0.2cm 0.0cm 0.2cm 0.1cm}, clip]{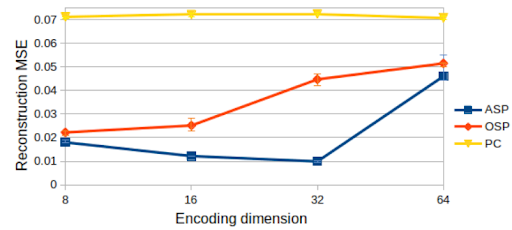}
	\caption{ASP maintained good performance as the encoding dimension increased, despite all models using 32 paired examples.}
	\label{fig:mnist1}
\end{figure}

We also analyzed how ASP, OSP, and PC performed as the encoding dimension changed.
A larger encoding dimension enables lower-error reconstructions in autoencoders, but mapping to and from a higher-dimensional space is more complex.
The results for a VAE of varying encoding dimension (all trained with 32 paired and 2,048 unpaired examples) are depicted in Figure~\ref{fig:mnist1}.

While PC and OSP tended to suffer as the latent dimension (and therefore social convention complexity) increased, ASP actually \textit{improved} initially before eventually degrading like the other models.
The improvement in performance from a latent dimension of 8 to 32 can be explained as ASP not suffering as much from the additional complexity while benefiting from the improved expressivity of higher-dimensional embeddings.

\subsection{CommNet Levers}

In our final domain, we adopted the communicative multi-agent architecture proposed by \citet{commnet} (CommNet) to use ASP in a lever-pulling game, played as follows:
three neural nets with shared weights were each given a unique integer agent ID from 0 to 9.
Based solely upon that ID and inter-net communication, the nets each simultaneously chose one of three levers.
Because they chose simultaneously, two nets could choose to pull the same lever; however, the reward (referred to as the ``supervised reward'' by the authors) was the proportion of levers for which the $i^{th}$ lever was pulled by the net with the $i^{th}$ ordered ID.
For example, if the nets were given IDs 8, 3, and 5 and pulled levers 0, 1, and 1, respectively, they received a score of $\frac{1}{3}$ because the net with the middle ID (5) pulled the middle lever (1), but the rest were out of order.

\begin{figure}[ht]
	\centering
	\includegraphics[width=\linewidth]{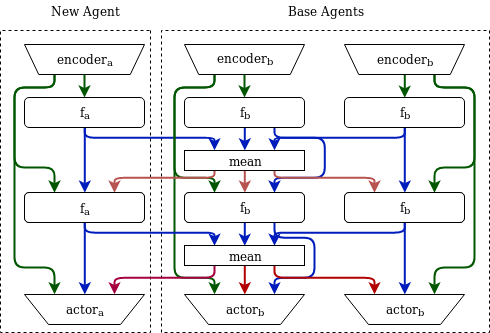}
	\caption{The architecture of a CommNet mixed team. In testing mixed teams, agent $a$ must coordinate across multiple steps with base agent $b$ despite having been trained in isolation.}
	\label{fig:commnet_arch}
\end{figure}

Using the neural architecture proposed in CommNet and training using epsilon-greedy deep Q learning, we trained a net to pull levers in the right order over 95\% of the time.
Figure~\ref{fig:commnet_arch} depicts a diagram of the mixed CommNet architecture.
The nets learned to coordinate by broadcasting information to each other about their agent ID and then sorting themselves.

Given the complex interactions enabled by the CommNet architecture, a new net must learn multiple social conventions in order to succeed.
We therefore introduced two adversaries when training with ASP: one for the outputs of each layer of f blocks (as shown in Figure~\ref{fig:commnet_arch}).
Theoretically, a single adversary could suffice, but we observed better results with different adversary instances.

\begin{figure}[ht]
	\centering
	\includegraphics[width=\linewidth, trim={0.2cm 0.5cm 0.2cm 0.1cm}, clip]{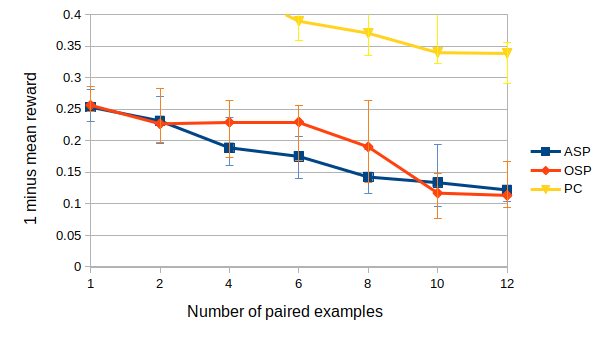}
	\caption{One minus reward in the lever task. ASP benefited most from a small amoutn of paired data.}
	\label{fig:commnet}
\end{figure}

ASP achieved high reward when given few paired examples.
We tested mixed teams by using one or two instances of a newly trained agent with two or one instances of the base agent (respectively).
In Figure~\ref{fig:commnet}, we demonstrate that the median error (defined as one minus the reward) of mixed teams over 20 trials decreased as the number of paired examples increased.
Both OSP and ASP outperformed PC by large margins, but ASP benefited more from a small number of examples.
As in the particle world domain, much of the variation in results came from differences among the different base models used in the trials: a Friedman's test on all results for 4, 6, and 8 paired examples was significant for ASP compared with OSP ($p < 0.05$) and for ASP compared with PC ($p < 0.01$).

\begin{figure}[ht]
	\centering
	\begin{subfigure}[b]{0.15\textwidth}
		\includegraphics[width=\textwidth, trim={2.3cm 1.5cm 2cm 1.8cm}, clip]{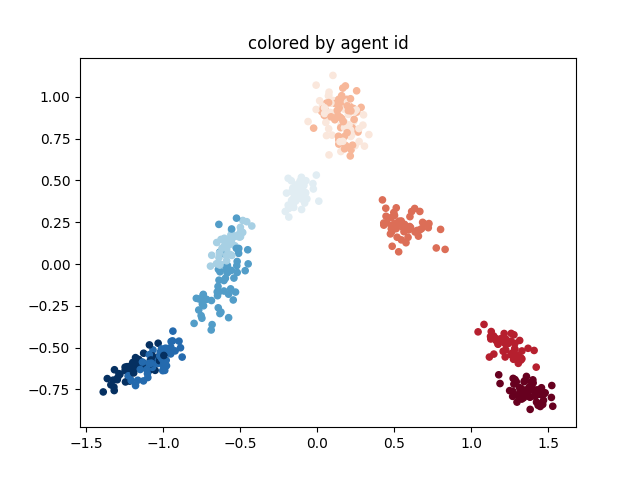}
		\caption{Base model}
	\end{subfigure}
	~ 
	\begin{subfigure}[b]{0.15\textwidth}
		\includegraphics[width=\textwidth, trim={2.3cm 1.7cm 2cm 1.8cm}, clip]{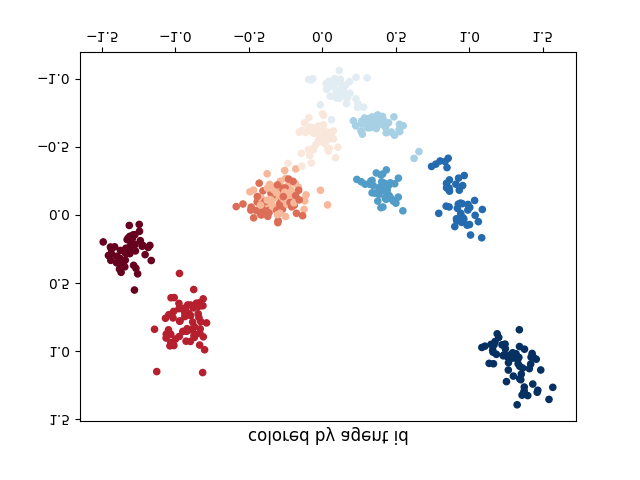}
		\caption{ASP + 0 paired}
	\end{subfigure}
	~ 
	\begin{subfigure}[b]{0.15\textwidth}
		\includegraphics[width=\textwidth, trim={2.3cm 1.7cm 2cm 1.8cm}, clip]{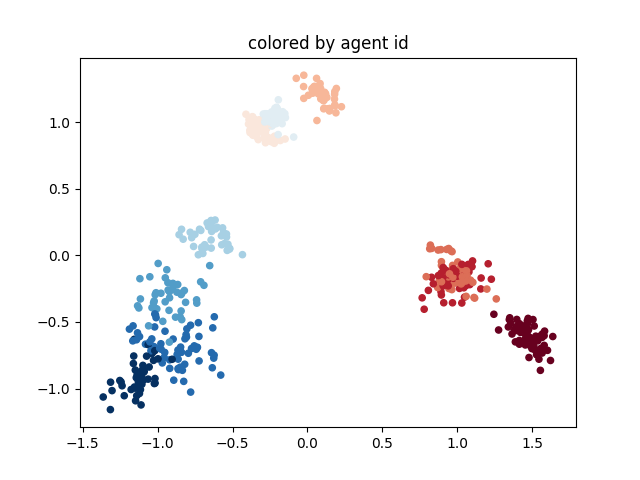}
		\caption{ASP + 8 paired}
	\end{subfigure}
	\caption{a) In a 2D projection colored by agent ID, communication vectors clustered; without paired data, ASP could not resolve ambiguity (b) until given paired data (c).}
	\label{fig:commnet_latents}
\end{figure}

As in analysis of the MNIST VAE, the communication vectors produced by CommNet are somewhat interpretable.
Figure~\ref{fig:commnet_latents} depicts the plot of the two-dimensional principal component analysis of the communication vectors produced in the last f module, colored by the agent ID fed into the encoder at the top of the net.
As reported in the original CommNet paper, the communication shows a pattern of sorting and clustering by agent ID.
ASP successfully recreated the base model's communication space, but needed paired examples to resolve the ambiguity of which agent ID corresponded with which communication cluster.

\section{Discussion and Future Directions}
Beyond domain-specific analysis, a broad pattern emerged across all experiments: for some number of paired examples in all tests, ASP produced agents that learned social conventions as well as agents trained with other techniques but with twice as much paired data.
This improvement is enabled through adversarial training in ASP that drives agents to match patterns in unpaired data.
While ASP does require an unpaired dataset, unpaired data are often far easier to gather than paired data.

Although we have demonstrated the utility of ASP in multiple domains, it is worth noting the technical and ethical concerns associated with its use.

First, using adversarial training with too little data may actually worsen performance by forcing a new model to exclusively employ a limited and sub-optimal set of actions.
Fortunately, such behavior is easily detectable in ASP: the adversary is fooled while self-play performance suffers.
We observed this behavior in early development and addressed it by decreasing the relative weight of the adversarial reward term and increasing the size of the unpaired dataset $U$.

Second, although this paper is based on the premise that we wish models to adopt existing social conventions, it is worth considering whether fully conforming to a social convention is always desirable.
In cooperative environments, it is easy to imagine how teams composed of heterogeneous agents, each with its own strength, might outperform homogeneous teams.
Furthermore, diversity in learning may allow a team to learn new strategies that might have been overlooked if all team members had adopted a uniform strategy early on.
Finally, and most perniciously, if adversarial techniques are used to force behaviors to resemble existing social conventions, a byproduct of training is a well-trained adversary.
Such an adversary, although potentially useful in cases such as fraud detection, could serve to target or penalize persons who do not behave like the majority.

\section{Contributions}
In this work, we have introduced ASP, an adversarial approach to learning social conventions with limited paired data.
Motivated by a French driver watching English traffic, we exploited patterns in unpaired data to lead new agents to the correct social convention.
ASP itself is broadly applicable: it works well in time-extended reinforcement domains, multi-agent coordination games, and even an autoencoder setting.

ASP's generality hints at possible directions for future work.
ASP could be used in human-robot teams to train robots to adapt to their human partners by merely observing them (gathering unpaired data) instead of asking how to behave in specific scenarios (gathering paired data).
In a similar vein, adversarial shaping of outputs could be used in interpretability research by encouraging representations to conform to understandable forms without explicitly prescribing all behavior.





\bibliographystyle{ACM-Reference-Format}  
\bibliography{sample-bibliography}  

\end{document}